\documentclass[runningheads]{llncs}
\usepackage{graphicx}
\usepackage{amsmath,amssymb} 
\usepackage{color}

\usepackage{booktabs}
\usepackage{enumitem }
\usepackage[linesnumbered,ruled,vlined]{algorithm2e}
\SetKwInput{kwSource}{Source}
\SetKwInput{kwProced}{Procedure}
\usepackage[misc]{ifsym}

\DeclareMathOperator*{\argmin}{arg\,min}
\usepackage{multirow}
\usepackage[flushleft]{threeparttable}
\usepackage[numbers,sort]{natbib}
\usepackage[hyphens]{url}

\begin{document}
\title{Target Aware Network Adaptation \\ for Efficient Representation Learning} 

\titlerunning{Target-aware Network Adaptation}
%
\author{Yang Zhong\inst{1,}$^{\textrm{\Letter}}$ \and
Vladimir Li\inst{1} \and
Ryuzo Okada\inst{2} \and
Atsuto Maki\inst{1}}
%
\authorrunning{Y. Zhong et al.}
%

\institute{KTH Royal Institute of Technology, Stockholm, Sweden 
\\ \email{\{yzhong, vlali, atsuto\}@kth.se} \\
\and
Toshiba Corporate Research and Development Center, Kawasaki, Japan\\
\email{ryuzo.okada@toshiba.co.jp}
}
\maketitle              
\begin{abstract}
This paper presents an automatic network adaptation method that finds a ConvNet structure well-suited to a given target task, e.g.\ image classification, 
for efficiency as well as accuracy in transfer learning.
We call the concept target-aware transfer learning.
Given only small-scale labeled data, and starting from an ImageNet pre-trained network, we exploit a scheme of removing its potential redundancy for the target task through iterative operations of filter-wise pruning and network optimization.
The basic motivation is that compact networks are on one hand more efficient and should also be more tolerant, being less complex, against the risk of overfitting which would hinder the generalization of learned representations in the context of transfer learning.
Further, unlike existing methods involving network simplification, we also let the scheme identify redundant portions across the entire network, which automatically results in a network structure adapted to the task at hand.
We achieve this with a few novel ideas: 
(i) cumulative sum of activation statistics for each layer, and
(ii) a priority evaluation of pruning across multiple layers.
Experimental results by the method on five datasets (Flower102, CUB200-2011, Dog120, MIT67, and Stanford40) show favorable accuracies over the related state-of-the-art techniques while enhancing the computational and storage efficiency of the transferred model.

\end{abstract}
%
%
\section{Introduction}

The methodology of constructing feature representations has been recently advanced from a well-known hand-crafted manner to a learning-based one.
Conventional hand-crafted features are typically designed by leveraging the domain knowledge of human experts \cite{LBP,surf,SIFT}. 
The learning based approaches often generate discriminative image representations using large-scale labeled datasets, such as ImageNet \cite{imagenet_cvpr09}, Places \cite{Place}, MS COCO \cite{MScoco}, and CelebA \cite{CelebA}, with deep and complex convolutional neural networks (ConvNets).
Thanks to the generic transferability, these learned deep representations from the ConvNets can also be utilized for other unseen tasks by means of transfer learning \cite{NIPS2014_5347,7328311,pmlr-v32-donahue14,Zeiler2014,Oquab14}:
one way is to directly use the pre-trained ConvNet to map images to the learned feature space with or without selecting more discriminative features \cite{zhong_icpr}; 
a more common practice, however, is to transfer a large off-the-shelf (OTS) model to a target task by mildly tuning the ConvNet parameters to make it more specific to a new target task \cite{CelebA,NIPS2014_5347,7328311}.

\begin{figure}[t!]
\includegraphics[width=0.7\textwidth]{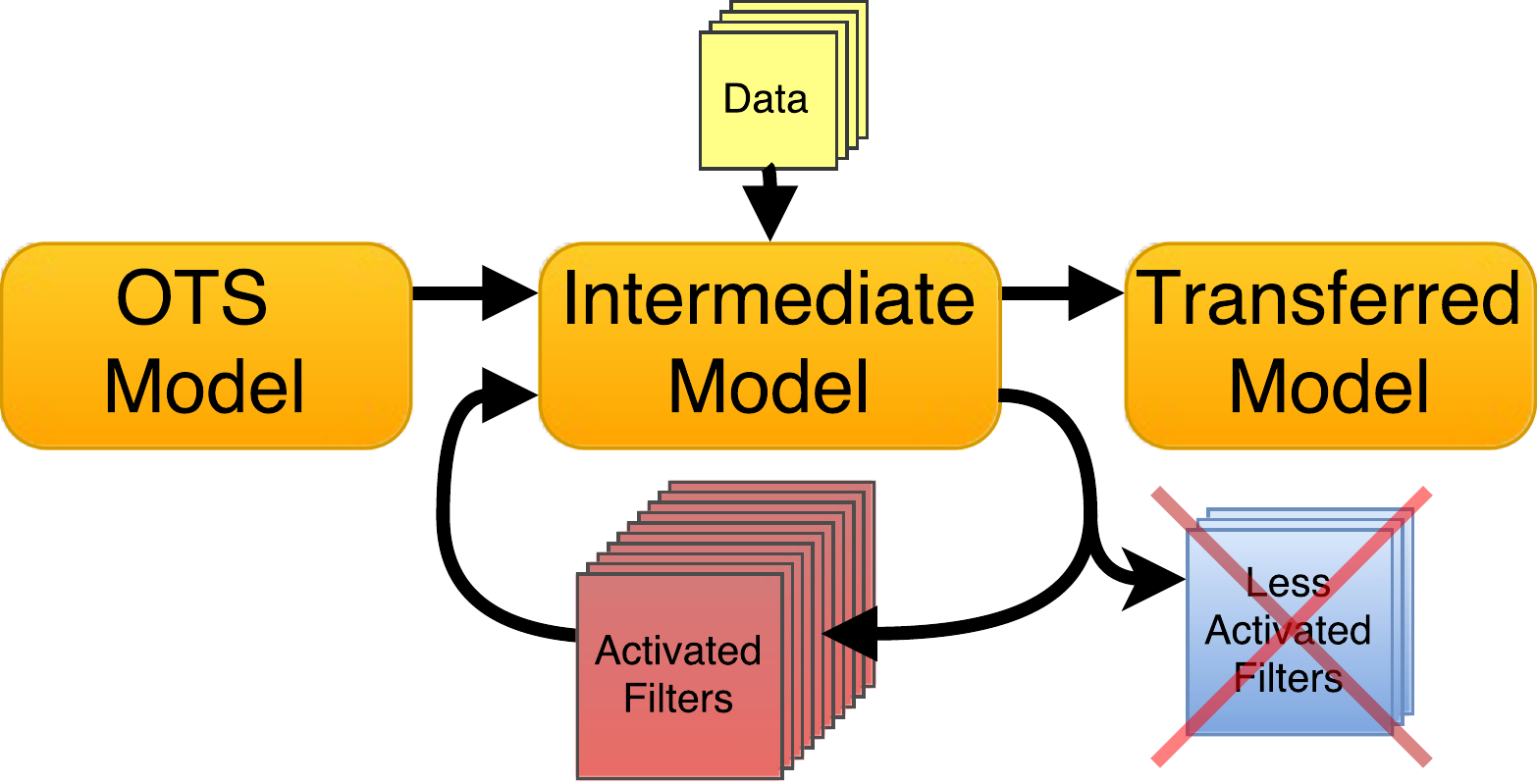}
\centering
\caption{A schematic of network adaptation proposed in this paper. An off-the-shelf (OTS) model is iteratively structured during the process of transfer learning where in each iteration, less significant filters in terms of activation statistics are discarded while the rest of the filters are reused in the next iteration.}
\vspace{-1em}
\end{figure}
       
Despite the successful applications of ConvNets along the scenario of transfer learning, a basic question with respect to the model structure is still unaddressed: a question about whether the off-the-shelf network is sufficiently complex or more than ample to model the target task. 
That is, most of the existing approaches to ConvNet transfer learning take a predefined network architecture as given and optimize the parameters therein, without seeking for a better suited network structure even though the given task can be much simpler and therefore requiring a less complex model.
Indeed, it has been shown that a model built for a source task could be less activated on target tasks \cite{DomainAdp_ICCV}.
This suggests that ConvNets can be made more compact to derive discriminative feature representations more efficiently in transfer learning scenarios.
Moreover, learning representations by reusing a model designed for a very large-scale problem (such as ImageNet classification \cite{imagenet_cvpr09}) for smaller target tasks would risk models to get overfitted to the target domain.
Tailoring a network suitable for a target task helps to enhance regularization when learning the target task representations.

As a natural progression for the transferred model to be more effectively adapted to target task, which we call {\it target-aware transfer learning}, an intuitive next step is to automatically remove possible redundancy from the transferred off-the-shelf model.
Besides an obvious benefit in improving computational efficiency, one could expect increased accuracy on a target task if it were possible to find a less complex model that is better structured to target data.
Thus, in this paper, we propose an automatic network adaptation method for target-aware transfer learning.
To this end, we exploit cumulative sum of activation statistics for each ConvNet layer to determine the priority of filters to prune across the network while fine-tuning the parameters in an iterative pipeline.
                                                                                                            
Although there are approaches to simplifying ConvNets in general \cite{Haoli} or learning a sparse network structure \cite{NIPS2016_6372,NIPS2016_6504,Zhou2016}, to the best of our knowledge this is the first work that addresses an automatic network adaptation for transfer learning. 
The work presented in this paper is close to \cite{EffiInf_ICLR} in that they deploy a framework consisting of iterative filter pruning and fine-tuning, and \cite{EffiInf_ICLR} also provides a comparative study on criteria for network pruning. 
As a development orthogonal to those criteria, our method has a functionality to modify and adapt the network structure according to the target task while also avoiding a greedy search to select parts of the network to be removed.
The contributions of this paper are summarized as follows:
\begin{itemize}
\item We propose a target-aware feature learning framework, {\it Network Adaptation},  which efficiently generates useful image representations. 
\item The Network Adaptation automatically tailors an off-the-shelf network (the weights and architecture) to the target task data at hand, so that the learned representations are more favorable and more efficient on the target tasks than the ones from simply fine-tuned off-the-shelf models.
\item The results highlight that the generalization of the features gets enhanced through the model compaction by our Network Adaptation for transferring given models to new target tasks.   
\end{itemize}

In Section \ref{sec:relate}, we introduce recent related work. 
The Network Adaptation method is described in detail in Section \ref{sec:algo}.
Comparative experimental results are demonstrated in Section \ref{sec:exp1} and Section \ref{sec:exp2} with ablation studies of different pruning strategies in Section \ref{sec:exp3}. 
Section \ref{sec:cl} concludes the paper.

\section{Related Work}
\label{sec:relate}

Transfer learning with deep ConvNets addresses a question of how to exploit a trained network for the sake of another task that has some commonality (to some unknown extent) to the source problem which the model has been trained on. 
For instance, one may use an ImageNet trained model and further train it to classify a subset of ImageNet classes, 
or even perform a rather different task \cite{posenet} which has less abundant data.
The goal of transfer learning is to construct feature representations for a target task.

A commonly adopted way of transferring a ConvNet is to fine-tune a source task model for a target task.
A small learning rate is often used to optimize the target-task orientated objective function so that the learned representations still preserve the generalization learned from the data in the source domain.
Feature selection could also be performed before fine-tuning, as different levels of features have different utility for the target task \cite{pmlr-v32-donahue14,Zeiler2014,Oquab14}.
One can see feature selection in this context as a plain data-dependent network compaction approach since some high-level layers may be skipped in a transferred network. 
In our work, network compaction is considered and applied on all across the network based on the activation statistics. 

Although ConvNets transfer learning has demonstrated higher performance than the conventional approaches in solving many computer vision problems \cite{7328311}, the enormous computational cost and memory footprint for using ConvNets have hindered applications in some practical scenarios.
To alleviate demanding hardware requirements, attentions have been paid to network compaction \cite{KD,deepcompress,thinet,DomainAdp_ICCV,EffiInf_ICLR}. 
Most of the existing compaction methods consider pruning models through coding \cite{deepcompress}, sparsity \cite{NIPS2016_6504,NIPS2015_5784}, matrix decomposition \cite{DomainAdp_ICCV}, or norm of filter weights \cite{Haoli}. 

Although it is known that small-scale target data often provides a poor sampling of the target domain, which causes overfitting with complex models, it is still very seldom that network compaction is explicitly employed to counteract overfitting to further help transfer learning.
Very recently, Pavlo et al.\ \cite{EffiInf_ICLR} proposed an iterative pruning method to optimize a fine-tuned networks for a target task. 
In every iteration, all the filters in the network are evaluated and one filter is pruned at a time based on saliency. 
The iterative loop needs to be carried out until a reasonable trade-off between accuracy and efficiency is reached.
In our method, it is sufficient to perform the network adaptation for much fewer iterations and it prunes many insignificant filters along the network in every single iteration.

On another aspect, to improve regularization for fine-tuning, it has been found that optimizing multiple helpful objectives leads to better model effectiveness compared to plain fine-tuning which only utilizes a cross-entropy loss \cite{Borrow,LwF}.
One can attempt to reduce the domain variance through certain metrics (i.e.\ perform domain adaptation) as in \cite{Tzeng_2015_ICCV,pmlr-v37-long15}.
It is also viable to perform multi-task learning explicitly.
In \cite{Borrow}, a data selection approach was developed to select (source-domain) data similar in low-level features to perform multi-task learning. 
Rather than relying on the availability of foreign data as in the aforementioned work, in \cite{LwF}, the predictions of target domain images given by the source model were recorded before training.
They were used later as replacements of extra training data.
Then, the current predictions were compared to the recorded ones to compose an additional objective in the loss function.
In this way, the network was able to be optimized on the target tasks, but at the same time be able to ``remember'' how to perform well on the source task. 
In this work, our approach also achieved better regularization without dependence on the availability of any extra data other than the target task at hand.

\section{Target-aware Network Adaptation}
\label{sec:algo}

As the intermediate representations along the network are likely to be redundant (over-completed) for a target task, it is reasonable to question if an off-the-shelf architecture is unnecessarily over-structured when being transferred to a target task.
By rethinking this commonly adopted fine-tuning procedure, in this section, we propose a network adaptation method to prune and structure an off-the-shelf network for a given target task in Section \ref{sec:algo1}. 

\subsection{Network Adaptation}
\label{sec:algo1}

\begin{algorithm}
\caption{Iterative Network Adaptation process.}
\label{algo:1}
\kwSource{1) An off-the-shelf network, 2) Labeled data in target domain.}
\BlankLine
 {\bf Step 0}: Fine-tuning the off-the-shelf network on the target task. \\
 \BlankLine
 
 \For{ iteration $i$ }{
  Break from the loop when $i$ reaches a certain value. 
  \BlankLine
  {\bf Step 1}: Prune filters in the network optimized in the previous step according to the activation statistics for the training data: 
  \begin{enumerate}[label=(\alph*)]
	\item On each layer, identify less significant filters in terms of cumulative sum of \\ average activations;
	\item Among all the network layers, prioritize the need for pruning identified \\ filters by the global priority.
  \end{enumerate}

  {\bf Step 2}: Fine-tune the pruned network on the target task, with the same objective function as in {\bf Step 0}. \\
  }
\end{algorithm}

The details of our network adaptation procedure are described in Algorithm \ref{algo:1}.
First, it takes an off-the-shelf network as a starting point and performs fine-tuning on the target task.
After that, in Step 1, it first collects activation statistics and prunes the trained network. 
Specifically, in Step 1(a), the average intensity of the activation maps is calculated on every layer after feeding in the entire training set. 

\begin{figure}[h]
\vspace{-1em}
\begin{center}
   \includegraphics[width=.65\linewidth]{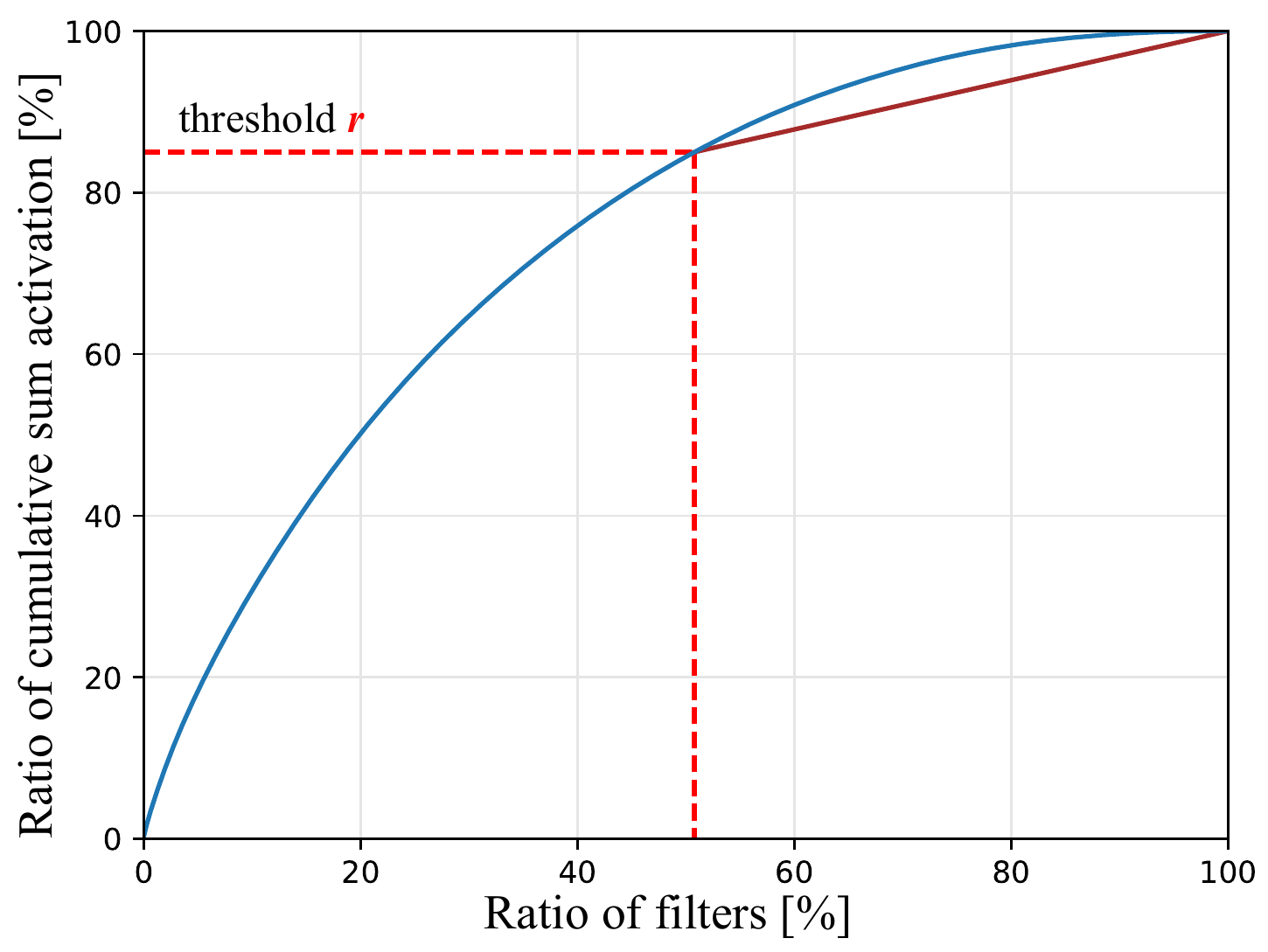}
\end{center}
\vspace{-1.5em}
   \caption{Cumulative sum of activations: an example from FC6 layer of the VGG-16 network based on the Flower102 dataset. The horizontal red broken line indicates the cumulative threshold ratio $r$, which is 85\% in this case. Filters corresponding to the right side of the vertical broken line are considered for pruning. The slope of the brown line represents the priority in Equation \ref{eqn:importance}.}
\label{fig:cumsum}
\end{figure}

Let us assume a convolutional layer which has an output activation tensor $A$, with a size of $H\times W \times K$ (where $K$ represents the number of output channels, and $H$ and $W$ stand for the hight and width of feature maps, respectively). 
The channel-wise activation, $\mathbf{a}^{(k)} (k=1,...,K)$, is simply calculated as: 
\begin{equation}
\mathbf{a}^{(k)} = \dfrac{1}{W\times H}\sum_{w=1}^W \sum_{h=1}^H A_{w,h}^{(k)}. 
\label{eqn:1}
\end{equation}
After feeding $N$ training images, $\mathbf{a}^{(k)}$ is averaged over $N$ instances: 
\begin{equation}
\mathbf{\bar{a}} = \{\bar{a}_{k}\}_{1}^{K}, \bar{a}_{k} = \sum_{n=1}^{N} (a_{n}^{(k)})/N.
\end{equation}
The channel-wise mean activation is then normalized by its $L1$ norm: 
\begin{equation}
\hat{\mathbf{a}} = \dfrac{\bar{\mathbf{a}}}{\Vert \bar{\mathbf{a}} \Vert_{1}}.
\end{equation}
Then, we compute the cumulative sum of descendingly sorted normalized mean activation $\hat{\mathbf{a}}$, which is denoted by 
\begin{equation}
\mathbf{c} = cumsum(\hat{\mathbf{a}}')\  | \ \hat{\mathbf{a}}' = sort(\hat{\mathbf{a}}).
\end{equation} 
Figure \ref{fig:cumsum} illustrates the curve $\mathbf{c}$.
Based on the cumulative sum, the filters corresponding to a cumulative sum value higher than a threshold  $r$ can be identified, i.e.,   
\begin{equation}
h = \argmin_k \mid c_k - r \mid \label{eqn:thresh}, 
\end{equation}

\begin{equation}
\mathbf{m}_p' = \{m_k'\}_{1}^K ,\  
m_k' =
\begin{cases}
1, k\le h\\
0, k > h
\end{cases}\\ ,
\end{equation}

\begin{equation}
\mathbf{m}_p = sort^{-1}(\mathbf{m}_p'),
\end{equation}
where $c_k$ is an individual element in $\mathbf{c}$ and $\mathbf{m}_p$ is a binary vector that indicates the indexes of potential filters to keep on a certain layer. 
The ratio between $h$ and $K$ is illustrated in Figure \ref{fig:cumsum} as the vertical broken line. 
The filters corresponding to the right side of that line are considered for pruning.

Next, in Step 1 (b), in order to further encourage the nature of data-dependent network adaptation, we additionally develop a way to calculate the priority. 
It helps us to decide whether to perform pruning or not to potential filters on a certain layer. 
The priority $s$ for layer $l$ is:
\begin{equation}
s^{(l)} = \dfrac{1-r}{1-h^{(l)}/K^{(l)}} \label{eqn:importance}
\end{equation}

Finally, the filters are pruned on the layers that have a priority lower than the mean of priority of all layers guided by $\mathbf{m}_p$;
pruning is therefore dynamically performed on the layers that are globally less important.
The pruned network is thereafter optimized in Step 2 and the network adaptation is performed iteratively.
The network adaptation can be terminated after a number of iterations when the validation accuracy starts to drop or the model size falls below a certain ratio.

\section{Experiments}
\label{sec:exp}
In this section, we first describe the datasets we used for the experiments and the details of how we trained the networks in Section \ref{sec:expset}. 
We then demonstrate the improvement brought about by our method in both accuracy and efficiency by comparing to the recent work addressing transfer learning and network compaction in an identical or a similar context in Section \ref{sec:exp1}. 
Section \ref{sec:exp2} and Section \ref{sec:exp3} reveal how the Network Adaptation impacts the network structure and computational efficiency. 
In addition, we also performed an ablation study which demonstrates the significance of activation-based network adaptation versus random filter pruning methods in Section \ref{sec:exp4}.   

\subsection{Experimental Setup}
\label{sec:expset}
In order to provide comprehensive results, our method is comparatively evaluated on five datasets: the Flower102 \cite{Flowers}, the CUB200-2011 \cite{WahCUB_200_2011}, the Dog120 \cite{DOG}, the MIT67 \cite{MITindoor}, and the Stanford40 \cite{Stanford40}, which represent classification tasks in different scenarios.
These datasets are among the most common benchmarks employed by the recent transfer learning related work.

\textbf{Flower102}
 has 8189 images of 102 flower classes.
 The training set and the validation set both have ten images respectively for each class.
The other images form the test set. 
In our experiments, we faithfully followed the training and test protocol, i.e.\ training only employs the 1020 images from the training set without any mixture with the validation instances.
\textbf{Dog120}
dataset contains 120 dog classes where each of the dog class has 100 training images and the test set contains 8580 images.
\textbf{CUB200-2011} (denoted by ``CUB200" in the following) 
dataset has 6000 training images and 5700 test images of 200 bird species.  
\textbf{MIT67} 
has 80 training images and 20 test images per indoor scene class. 
\textbf{Stanford40} 
dataset contains images of human actions of 40 classes, each of which contains 180 to 300 images. 
On the Dog120, CUB200, MIT67, and Stanford40, $10\%$ training images were separated to form the validation set for model training. 

To augment the training images, we employed random jittering and subtle scaling and rotation perturbations to the training images.
We resized images of all involved datasets to $250 \times 250$ and the aspect ratio of the images was retained by applying zero padding all the time.
During test time, we averaged over the network responses from the target-task classifiers over ten crops which were sampled from the corners and the centers of originals and the flipped counterparts.

There are a few well-known off-the-shelf networks that can be used in this study such as the AlexNet \cite{AlexNet}, the InceptionNet \cite{goingdeeper}, the ResNet \cite{Resnet}, and the VGGNet \cite{VGG}.
We choose the VGG-16 architecture in our experiments as it is a well studied and widely utilized structure for different kinds of computer vision problems.
More importantly, it facilitates comparisons to be as fair as possible since the VGG-16 performs neutrally comparing to the ``main-stream'' architectures. 
The off-the-shelf VGG-16 model used in our experiments was pre-trained on the ImageNet \cite{imagenet_cvpr09} and is publicly accessible \cite{OTS_model}.

For training networks, we used a batch size of 32 with a conservative learning rate of $10^{-4}$, which was helpful to achieve stable convergence, on all the datasets.
The learning rate was dropped by a factor of 10 once the validation loss stopped decreasing.
On most of the datasets, the learning rate was reduced to $10^{-5}$ within 20 epochs. 
The model training was terminated when the validation loss stopped decreasing. 
The model snapshots which performed best on the validation sets during the training were used for performance evaluation on the corresponding test sets. 
We set weight decay to 0.0005 and Dropout ratio to 0.5.  

\subsection{Comparative Performance Evaluations}
\label{sec:exp1}
Our experiments in this section focus on comparing our Network Adaptation approach with the standard fine-tuning on the selected target tasks.
One straightforward way to apply the Network Adaptation to an off-the-shelf architecture is to perform filter-wise pruning along the entire network, i.e.\ pruning filters from Conv1\_1 until FC7 in the VGG-16 architecture. 
Considering the ``blessing of dimensionality'' \cite{blessing}, however, another rational choice is to apply the Network Adaptation excluding the FC7 layer. 
In this case, the dimensionality of the high-level feature is maintained, which could be helpful to ensure the discrimination power, although at a cost of marginal computational overhead. 
In the following experiments, we evaluate both options by running independent Network Adaptation processes from the same fine-tuned starting point (i.e.\ from an identical model at ``Step 0'' in Algorithm \ref{algo:1}  on each dataset).

Specifically, in our experiments, we first performed fine-tuning on each dataset and then performed Network Adaptation iteratively. 
At each iteration, we used the best performing model on the validation set for performance evaluations. 
It was then used as the starting model standpoint for the next step. 
For the Network Adaptation, it is easy to see that the validation accuracy could fluctuate in the first few iterations and decrease eventually. 
With this regard, we ran 20 Network Adaptation iterations and selected the best performing model (on the validation sets) for the performance evaluation and compared them to the corresponding fine-tuning baselines.

\begin{table}[h]
\centering
\caption{Comparing the test accuracy of fine-tuning with Network Adaptation (NwA) of different design options. 
The test accuracy of Network Adaptation given here was determined by the best validation accuracy on the corresponding dataset. The Network Adaptation iteration (shortened to ``Iter.'') number when the best validation accuracy occurred is listed in the parenthesis after test accuracy. Threshold ratio $r$ was set to $2\%$.}
\label{tab:testacc}
\begin{tabular}{cccccc}
\toprule
           & Fine-tune	& 	& NwA w/FC7 		 			&	& NwA w/o FC7 			\\ \midrule
CUB200     & 75.84\%   	&	& 76.74\% (@Iter. 1)  			& 	& \textbf{77.49}\%  (@Iter. 2)    \\ \hline
Dog120     & 82.79\%   	&	& \textbf{82.88}\% (@Iter. 1)	&	& 82.73\%  (@Iter. 2)    	\\ \hline
Flower102  & 84.96\%   	&	& \textbf{85.14}\% (@Iter. 1)  	& 	& 83.51\%  (@Iter. 9)    \\ \hline
MIT67      & 70.30\%   	&	& \textbf{71.34}\% (@Iter. 5)  	& 	& 71.34\%  (@Iter. 4)    \\ \hline
Stanford40 & 76.23\%   	&	& \textbf{77.19}\% (@Iter. 5)  	& 	& 77.01\%  (@Iter. 5)    \\ 
\bottomrule
\end{tabular}
\end{table}

Table \ref{tab:testacc} shows how the Network Adaptation performs over the standard fine-tuning. 
First, it can be seen that by keeping the dimensionality of FC7 (NwA w/o FC), it outperforms the standard fine-tuning except on Flower102 and Dog120 dataset. 
When the dimensionality of the FC7 layer is reduced, Network Adaptation outperforms the standard fine-tuning on all the datasets by an error rate reduction of 3.72\% on CUB200, 0.52\% on Dog120, 1.20\% on Flower102, 3.50\% on MIT67, and 4.03\% on Stanford40. 
Remember that such a performance gain was achieved by using more compact networks and only the target task training data.
It suggests that how to restructure the network structure to achieve better model effectiveness could also be taken into consideration when transferring an off-the-shelf model to a new target task. 

Second, by applying the Network Adaptation along the entire VGG-16 architecture, even better test accuracy can be achieved on almost all datasets except CUB200. 
This means that restructuring the entire network not only results in even lower computational cost but also promises better performance margin on average. 
In other words, keeping the dimensionality of the high-level abstractions may not be as important in general.
In the following experiments, we perform the Network Adaptation along the entire network unless otherwise stated.

To comprehensively evaluate the Network Adaptation, we focus on the recent approaches which employed network compaction based methods (but brought about accuracy gains) to address transfer learning problems.
Given that regularization may be improved by building compact networks for target tasks, a recent state-of-the-art method \cite{LwF} that explicitly enhanced regularization through multi-task learning is also compared to our approach. 
These approaches may not employ exactly the same network architecture and other experimental setups (e.g.\ \cite{DomainAdp_ICCV} used a deeper VGG-19 network) as ours, which consequently resulted in slightly different fine-tuning baseline accuracy.
To handle the discrepancies, a reasonable comparative evaluation is to compare the relative accuracy gain of these methods, as shown in Table \ref{tab:comp}.
 
\begin{table}[h]
\centering
\caption{Comparing the gain of the test accuracy on various datasets with the recent related approaches. 
Each entry is the accuracy gain compared to the corresponding fine-tuning baseline. Threshold ratio $r$ set to $2\%$ in our Network Adaptation.}
\begin{threeparttable}
\begin{tabular}{ccccccccccc}
\toprule
        		            						& CUB200 &	& Dog120 &	&	Flower102	&	& MIT67		&	& Stanford40 \\ \midrule
Best of Deep Compression \cite{DomainAdp_ICCV} 		& 0.12\% &	& ---	 &	&  0.1\%       	&	& ---   	&	& 0.79\%     \\ \hline
Best of Efficient Inference \cite{EffiInf_ICLR}		& 0.5\%  &	& ---    &	&  ---      	&	& ---      	&	& ---       \\ \hline
LwF \cite{LwF}         								& -0.6\% &	&  ---   &	&   ---			&	& 0.3\%  	&	&  ---       \\ \midrule
Ours, w/o FC7       								& 1.65\% &	&-0.04\% &	&  -1.45\%    	&	& 1.04\% 	&	& 0.78\%	 \\ \hline
Ours, w/ FC7        								&  0.9\% &	&0.09\%  &	&  0.18\%	  	&	& 1.04\% 	&	& 0.96\%      \\ \bottomrule
\end{tabular}
    \begin{tablenotes}
      \small
      \item Note that our total compression ratio is on-par with that in \cite{DomainAdp_ICCV}, see Figure \ref{fig:red}. The best gain on CUB200 from \cite{EffiInf_ICLR} was manually retrieved from Figure 4 in \cite{EffiInf_ICLR}.
    \end{tablenotes}
\end{threeparttable}
\label{tab:comp}
\vspace{-1em}
\end{table}

On the effectiveness of the learned representations, it is clear that the Network Adaptation contributed to superior accuracy than other methods except for the case when it was not applied to the FC7 layer of a VGG-16 network on Flower102.
We argue that this is an obvious advantage to some existing network pruning methods which have to make compromises to the performance on target tasks.  
The accuracy gain, in general, is attributed to the improved regularization (which was achieved in different ways in these methods).
It can be seen that making networks more compact is potentially a useful means to improve regularization owing to the positive gains. 
Our Network Adaptation indeed achieved better regularization than others given the more favorable performance improvements.
In addition, we noticed that the network compaction process is quite efficient with the Network Adaptation approach as it does not depend on exhaustive search as in \cite{EffiInf_ICLR}; the compaction is also more effective for reducing the computational cost than in \cite{DomainAdp_ICCV} as filters across the entire network were simultaneously considered for pruning.
The network compaction effects of our approach are demonstrated with details in the following sections.

\subsection{Evolution of Network Size during Network Adaptation}
\label{sec:exp2}

The Network Adaptation, on the one hand, can be used to improve the model effectiveness on the target tasks,  on the other hand, it also enforces target data-driven model compaction which effectively reduces the size of ConvNets.
In this way, one can make a trade-off between the network size and the model accuracy on the target task with our Network Adaptation. 

\begin{figure}[h]
\begin{center}
   \includegraphics[width=1\linewidth]{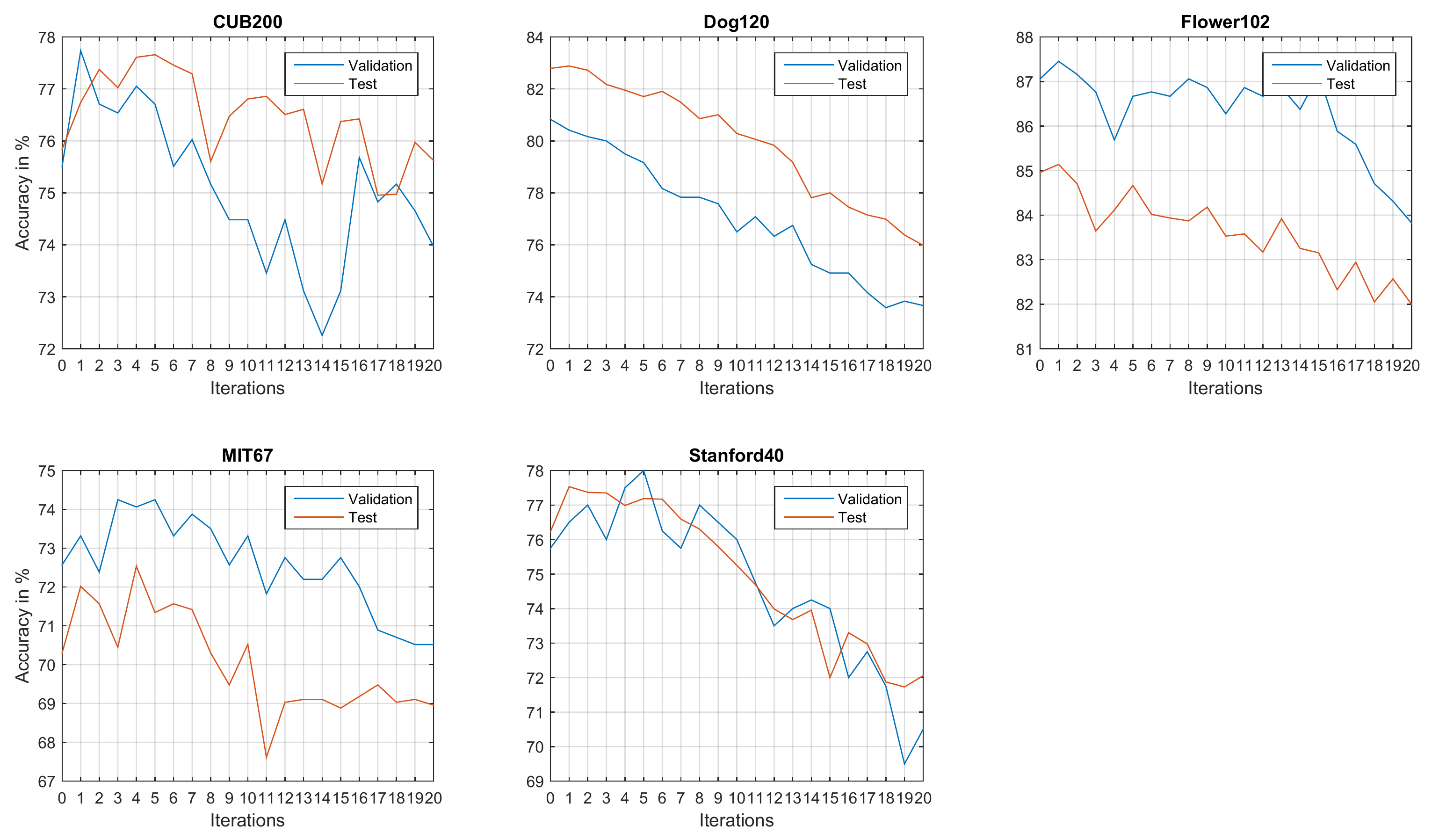}
\end{center}
\vspace{-2em}
   \caption{Validation accuracy versus test accuracy on the datasets during 20 iterations of Network Adaptation. 
}
   \label{fig:valtest}
\vspace{-1em}
\end{figure}

To demonstrate how the discriminability of the learned representations varies on the target tasks along the Network Adaptation process, we show the validation accuracy and the test accuracy along 20 Network Adaptation iterations in Figure \ref{fig:valtest}.
One can find that after fine-tuning on the target tasks (Iteration 0) the Network Adaptation is able to boost the discrimination power of the learned representations, which was reflected on the validation and test accuracy in the first few iterations.
It is clear that the improvement in the test accuracy is data dependent.
We can expect around or more than 1\% accuracy increment on MIT67, CUB200, and Stanford40, which have moderately large training sets.
But the improvement was less significant when training data was too sparse or much too large.
As on the Flower102, it had only 10 training images per class, while on Dog120 each class had around 100 images for training for each class, which was 2 to 3 times larger than other datasets. 
The accuracies dropped at later iterations when the networks lost more capacity.
As our threshold $r$ was only 2\%, the test accuracies were not significantly decreased even at Iteration 20.

\begin{figure}[h]
\begin{center}
   \includegraphics[width=1\linewidth]{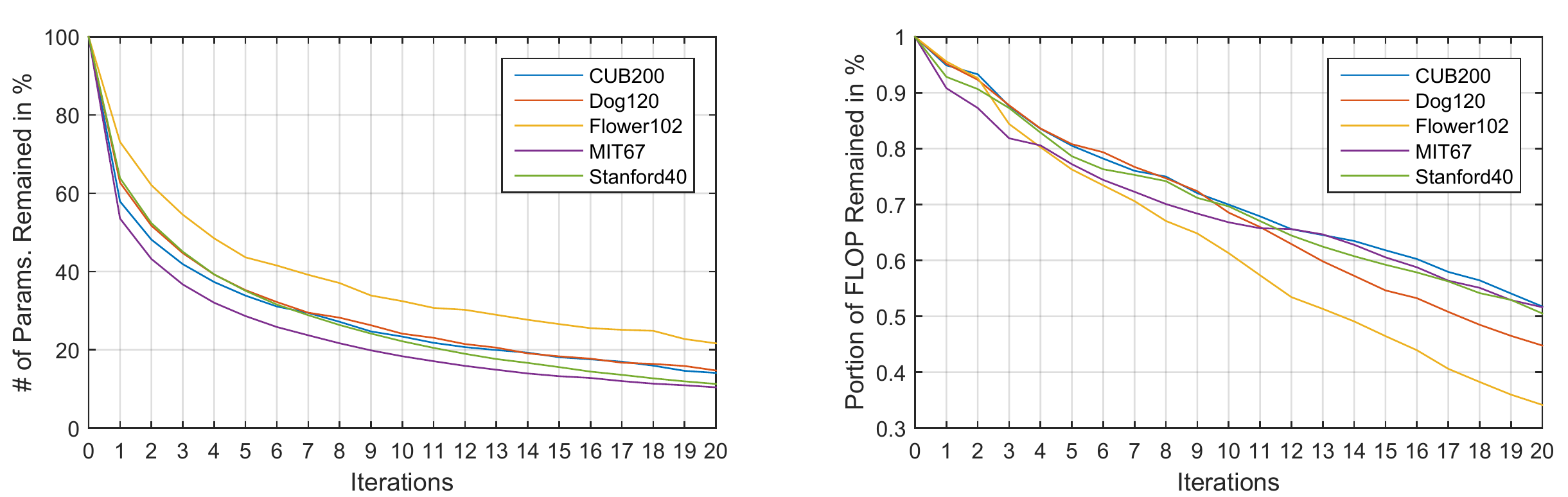}
\end{center}
\vspace{-1em}
   \caption{Normalized reduction of the number of network coefficients (left) and the corresponding normalized reduction of FLOP (right) with a VGG-16 architecture achieved by the Network Adaptation on different datasets.}
   \label{fig:red}
\vspace{-.5em}
\end{figure}

The reduction of the total number of parameters and the corresponding computational costs at each Network Adaptation iteration are shown in Figure \ref{fig:red}. 
First, it can be seen that the total number of network coefficient was reduced in an exponential trend.
The network size can be halved in the first five iterations on all the tasks even with an insignificant threshold value. 
The network size got linearly reduced after Iteration 8. 
Second, the Network Adaptation resulted in roughly a linear decrementing trend in the computational cost in terms of Floating-point Operation (FLOP).
For both of the decreasing curves of network size and FLOP, a similar data-dependent character can be observed; on the tasks with more abundant data, they shared a common tendency which was deviated on the task with the least data.

\begin{figure}[]
\begin{center}
   \includegraphics[width=.95\linewidth]{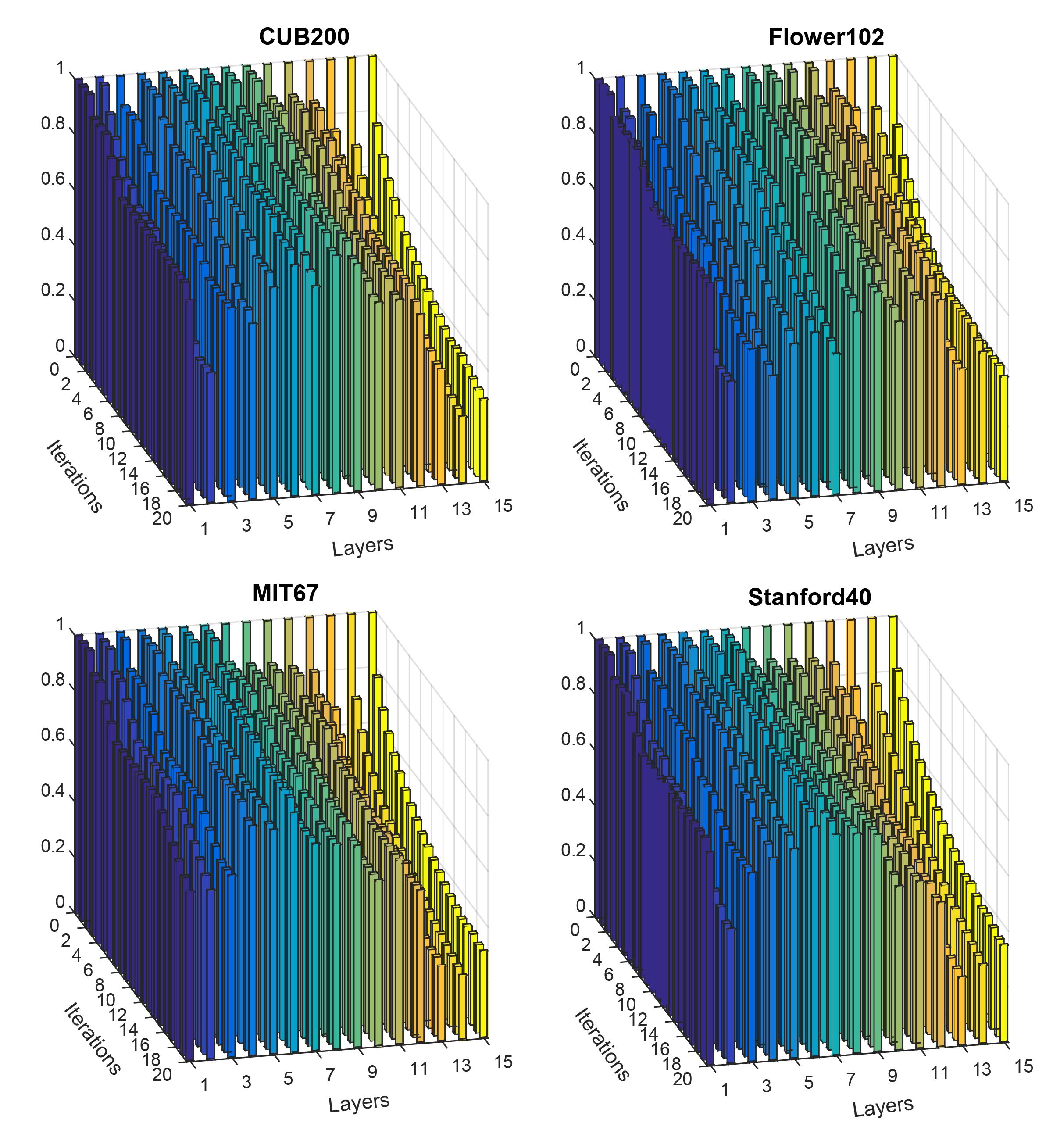}
\end{center}
\vspace{-2em}
   \caption{Relative change of layer width at each Network Adaptation iteration. The ``Layers'' scale, from 1 to 15, represents the learnable layers from ``Conv1\_1'' till ``FC7'' in a VGG-16 structure.}
   \label{fig:3d}
\vspace{-1em}
\end{figure}

The detailed sizes of each layer width (number of filters) for the 20 Network Adaptation iterations on the selected datasets are visualized in Figure \ref{fig:3d}.
It is easy to see that a VGG-16 architecture was (re-)shaped differently by our Network Adaptation on different tasks. 
An interesting phenomenon is that the lower level filters in the vicinity of Conv2\_1 layer were pruned relatively more than other layers, but on MIT67 the pruning rate was not as large as on the others.
This suggests a large but task-specific redundancy in the lower layers of the off-the-shelf model.
The same situation can be observed on higher-level convolutional layers.
It can, therefore, be inferred that the learned high-level image representation at the FC7 layer indeed depends on different combinations of the convolutional features:
the indoor scene classification MIT67 relies  more on lower level features than other tasks;
CUB200 and Stanford40 require more descriptive mid-level representations;
Flower102  more or less relies on higher-level convolutional features.

\subsection{Impact of Threshold Ratio}
\label{sec:exp3}

In this section, we evaluate the impact of threshold ratio ($r$) to the effectiveness of Network Adaptation. 
Specifically, we perform ten iterations of Network Adaptation with 2\%, 5\%, and 10\% threshold ratios and evaluate the best models indicated by the validation sets to explore the performance of the Network Adaptation with different setups.

\begin{figure}[h]
\vspace{-.5em}
\begin{center}
   \includegraphics[width=1\linewidth]{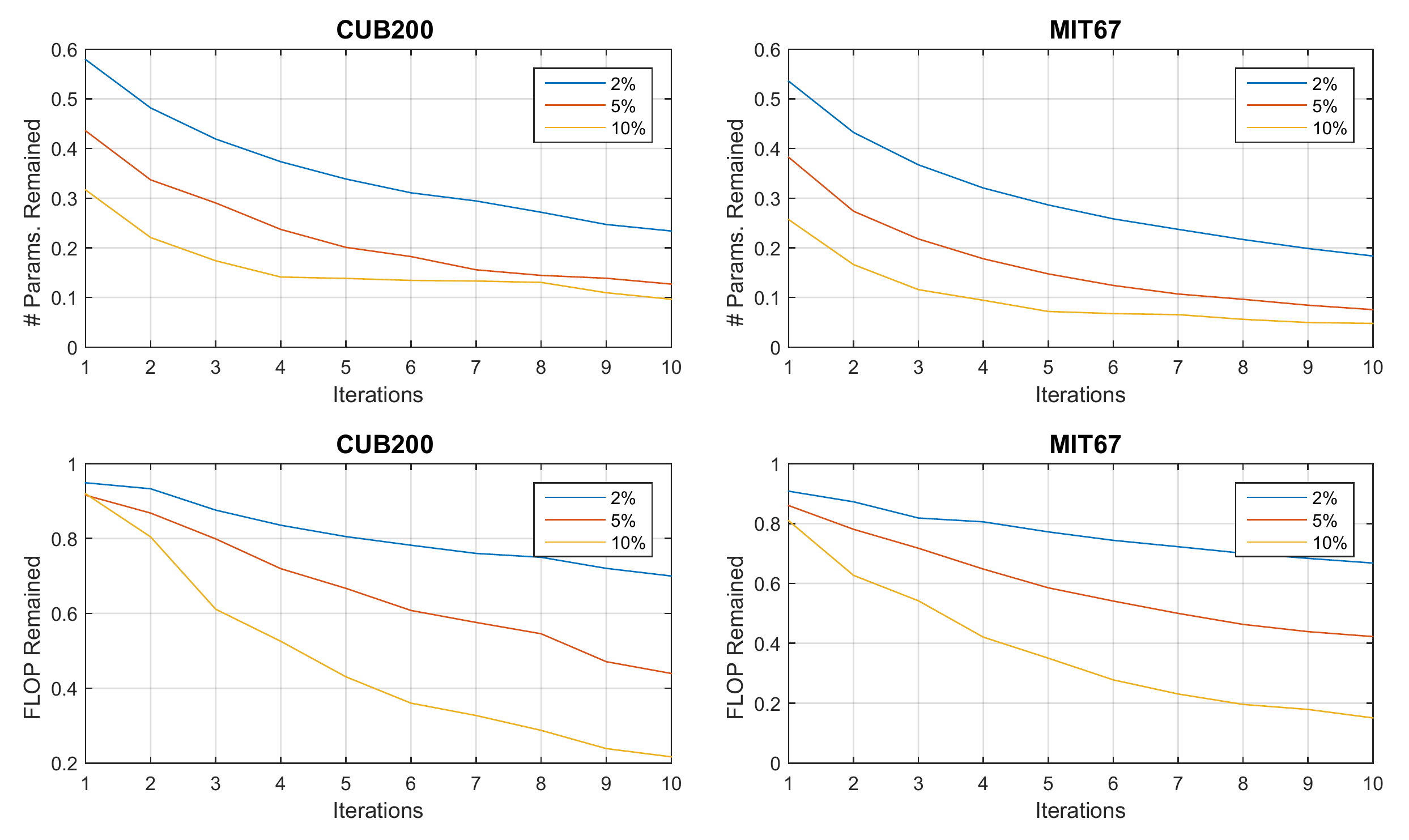}
\end{center}
\vspace{-1.5em}
   \caption{Comparing pruning effects by using different threshold ratios in Network Adaptation. The x-axis represents the iterations of the  Network Adaptation from 1 to 10, and the relative residual portion of network parameters and computational costs of a VGG-16 architecture is shown in y-axis.  }
\label{fig:cmprates}
\end{figure}

As shown in Figure \ref{fig:cmprates}, by employing different threshold ratios in Network Adaptation, a target VGG-16 model can be compacted to different extents; the higher the rate is, the more the models get compact in general.
For a single threshold rate, however, a VGG-16 model can become compact to different levels as well.
This is due to fact that the Network Adaptation is totally a target data guided method. 
E.g., for a threshold ratio of 10\% at Iteration 10, more than 10\% of the coefficients were retained on the CUB200, but on the MIT67 the remaining portion was only around 6\%. 
MIT67 data yielded stronger pruning effect in the Network Adaptation process.

\begin{table}[h]
\vspace{-.5em}
\centering
\caption{Test accuracy on the CUB200 and the MIT67 with different threshold ratios. The Network Adaptation iteration which gave the best performing model (on validation set) are given in parenthesis.}
\label{tab:aa}
\begin{tabular}{cccc}
\toprule
\multirow{2}{*}{} & \multicolumn{3}{c}{Threshold Ratios}                        \\ \cline{2-4} 
                  & 2\%               & 5\%               & 10\%              \\ \hline
CUB200            & 77.49\% (@Iter. 2) & 76.19\% (@Iter. 4) & 76.64\% (@Iter. 1) \\ \hline
MIT67             & 71.34\% (@Iter. 4) & 71.49\% (@Iter. 2) & 69.48\% (@Iter. 2) \\ \bottomrule
\end{tabular}
\vspace{-.5em}
\end{table}

Table \ref{tab:aa} lists the test accuracy values on the CUB200 and the MIT67 under three threshold ratios. 
It reveals a rough relation between the number of iterations and the threshold ratios with respect to the best test accuracy; a mild pruning rate may require more  steps to achieve the best accuracy and vice versa. 
From these results, we may infer that a threshold ratio between 5\% and 10\% would be a practical choice which balances between the training speed and the model accuracy. 

\subsection{Significance of Least Activation Pruning}
\label{sec:exp4}
In our Network Adaptation, we prune filters which correspond to the least activations of a given dataset. 
As shown in Figure \ref{fig:3d}, it shapes the model architecture based on the target data in a way that it removes different numbers of filters on different layers.  
To highlight the importance of this pruning strategy, we evaluate two alternative filter pruning methods which can be potentially used by our Network Adaptation framework in two stages. 

In the first experiment, we compared the performance discrepancy between removing a random 10\% of filters versus pruning 10\% of the least activated filters in each layer along the whole network.
In both ways, the models were uniformly pruned along the structure but these pruning strategies cannot (re-)shape the model structure.
Due to the randomness, we ran five independent experiments for one iteration with both strategies and compared the average accuracy.

The second experiment intended to further verify that randomly pruning filters performs less effectively than filters selected based on activation. 
This is done by directly comparing random pruning versus activation based pruning in the Network Adaptation procedure, where random pruning removes an equal amount of filters as by the Network Adaptation in each layer in each iteration (we also ran 20 iterations with the threshold ratio $r$ set to 2\%).
The same starting models at ``Step 0'' as in the previous experiments were used in these experiments but the Network Adaptation iterations were run independently for the comparative studies here.  
The results are listed in Table \ref{tab:abs1}.

\begin{table}[htbp]
\vspace{-.5em}
\centering
\caption{Comparing other pruning strategies to the activation based approach we used in the Network Adaptation. ``Least Act.'' indicates ``Least Activation''. ``Random'' is short for random pruning strategies for each experiment. }
\begin{tabular}{cccccc}
\toprule
       & \multicolumn{2}{c}{Experiment 1}                  &                  & \multicolumn{2}{c}{Experiment 2}                                 \\ \hline
       & Random 10\% 	& Least Act., 10\%   & \ & Random             & NwA                                         \\ \hline
CUB200 & 76.02\%      	& \textbf{76.46}\% 		& \ & 75.96\% (@Iter. 1) & \textbf{77.03}\% (@Iter. 7) \\ \hline
MIT67  & 65.33\%      	& \textbf{67.60}\% 		& \ & 68.73\% (@Iter. 4) & \textbf{71.34}\% (@Iter. 5) \\ \bottomrule
\end{tabular}
\label{tab:abs1}
\vspace{-.5em}
\end{table}

The results of Experiment 1 show that uniformly pruning a network performs inferior to the Network Adaptation.
Comparing the accuracy of the random pruning strategy in both experiments to that of the Network Adaptation, one can see that in general random filter pruning performs less effectively than activation based as used in our method, i.e. activation based methods offer a more reasonable pruning.

\section{Conclusions}
\label{sec:cl}

We proposed an iterative network adaptation approach to learn useful representations for  transfer learning problems which have access to only small-scale target data.
Our approach automatically structures model architecture by pruning less important filters, in which a data dependent method is developed to identify less important filters. 
Being adapted in the proposed pipeline, the transferred models are shown to be more computationally efficient and demonstrate at least comparable performance to the widely used ``fine-tuning'' practice and even the related state-of-the-art approaches.

In addition to the experimental results, the current development of our Network Adaptation approach
left us some interesting open questions.
First, for transfer learning, how to enhance regularization with limited labeled data is not fully studied yet;
we demonstrated that model compaction would be a promising candidate, but it will be also interesting to evaluate other alternatives.
Second, for network adaptation, it is ideal to have equally favorable performance to all target tasks but the classification accuracies by our approach slightly decreased in some situations. 
This hints us that it is also helpful to consider partly expanding the network architecture according to data to complement the capacity of the transferred model.

\section*{Acknowledgements}
We acknowledge fruitful discussions and comments to this work from colleagues of Toshiba Corporate Research and Development Center. 
We thank NVIDIA Corporation for their generous donation of NVIDIA GPUs.


%
%
\bibliographystyle{splncs04}
\bibliography{egbib}
%




\end{document}